\documentclass{article}

\usepackage[preprint]{neurips_2022}

\usepackage[utf8]{inputenc} 
\usepackage[T1]{fontenc}    
\usepackage{hyperref}       
\usepackage{url}            
\usepackage{booktabs}       
\usepackage{amsfonts}       
\usepackage{nicefrac}       
\usepackage{microtype}      
\usepackage{xcolor}         
\usepackage{amsmath, amsfonts, amsthm, amssymb}
\usepackage{algorithm, algpseudocode}
\usepackage{graphicx}
\graphicspath{ {./img/} }
\DeclareMathOperator*{\argmax}{arg\,max}
\DeclareMathOperator*{\argmin}{arg\,min}

\title{k-Means Maximum Entropy Exploration}
\author{%
  Alexander Nedergaard \\
  Institute of Neuroinformatics \\
  ETH Z\"urich\\
  \texttt{anederga@ethz.ch} 
  \And
  Matthew Cook \\
  Institute of Neuroinformatics \\
  ETH Z\"urich\\
  \texttt{mc@ethz.ch} 
}

\begin{document}

\maketitle

\begin{abstract}
Exploration in high-dimensional, continuous spaces with sparse rewards is an open problem in reinforcement learning. 
Artificial curiosity algorithms address this by creating rewards that lead to exploration.
Given a reinforcement learning algorithm capable of maximizing rewards, the problem reduces to finding an optimization objective consistent with exploration. 
Maximum entropy exploration uses the entropy of the state visitation distribution as such an objective. However, efficiently estimating the entropy of the state visitation distribution is challenging in high-dimensional, continuous spaces. 
We introduce an artificial curiosity algorithm based on lower bounding an approximation to the entropy of the state visitation distribution. The bound relies on a result we prove for non-parametric density estimation in arbitrary dimensions using k-means. We show that our approach is both computationally efficient and competitive on benchmarks for exploration in high-dimensional, continuous spaces, especially on tasks where reinforcement learning algorithms are unable to find rewards. 
\end{abstract}
\section{Introduction}
Reinforcement learning (RL) algorithms learn to take actions through interaction with their environment, adapting their behavior to maximize rewards. In recent years, RL algorithms have surpassed humans in several challenging domains such as Atari \citep{mnih2015}, Go \citep{silver2016}, Dota \citep{berner2019} and Starcraft \citep{vinyals2019}. Many real-world problems can be tackled by RL, however these problems often involve high-dimensional, continuous spaces and sparse rewards. Exploration, the problem of finding rewards, remains a major obstacle to practical applications of RL. Count-based exploration theory \citep{strehl2005}\citep{kolter2009}\citep{bellamare2016} has focused on the exploration-exploitation trade-off, ignoring an important aspect of real-world exploration: Sparse rewards are very difficult to find in high-dimensional, continuous spaces and RL algorithms cannot learn without rewards.

Artificial curiosity \citep{schmidhuber1991} is an approach to exploration where intrinsic rewards are generated and maximized. Given an RL algorithm capable of maximizing rewards, exploration reduces to generating rewards that lead to exploratory behavior when maximized.   
The potential capabilities of artificial curiosity algorithms are open-ended as these algorithms not limited by predefined tasks and thus constitute a candidate path toward artificial general intelligence.    
Several competitive approaches to exploration maximize intrinsic rewards related to state-transition surprise \citep{pathak2017} or state novelty \citep{burda2018}. These approaches are usually heuristically explained as information- or novelty seeking, lacking a clear mathematical basis for why they work. We have intutions about what it means to explore, but no clear criterion for ranking exploration in the absense of extrinsic rewards.

Maximum entropy exploration \citep{hazan2019} proposes to maximize the entropy of the state visitation distribution for exploration. The approach provides a mathematical basis for exploration based on the heuristic that visiting states uniformly is the best we can do without extrinsic rewards. Combining artificial curiosity and maximum entropy exploration provides a promising and principled approach to exploration. However, it requires efficiently computing the entropy of the state visitation distribution, which is not possible in high-dimensional, continuous spaces using standard non-parametric density estimation methods such as kernel density estimation or Gaussian Mixture Models. 

Here, we introduce \textbf{k}-Means \textbf{M}aximum Entropy \textbf{E}xploration (KME), an approach to maximum entropy exploration that uses a variant of k-means to estimate entropy. We prove a result for non-parametric density estimation in arbitrary dimensions and use this result to derive an approximate lower bound on entropy. We then construct an artificial curiosity algorithm that efficiently computes intrinsic rewards based on this bound. 
Our experiments show that KME is both computationally efficient and competitive on benchmarks for exploration in high-dimensional, continuous spaces, especially on tasks where RL algorithms are unable to find rewards.
\section{Related work}
Several maximum entropy exploration methods have emerged since the introduction by \cite{hazan2019}.  
State Marginal Matching \citep{lee2020} considers maximum entropy as a special case of shaping the state visitation distribution and Geometric Entropic Maximization \citep{guo2021} maximizes a geometric variant of entropy. 
Most methods use a k-nearest neighbor entropy estimate, including Maximum Entropy POLicy optimization (MEPOL) \citep{mutti2021}, Active Pre-Training (APT) \citep{liu2021a}, Active Pre-Training with Sucessor Features (APS) \citep{liu2021b}, Proto-RL \citep{yarats2021} and Random Encoders for Efficient Exploration (RE3) \citep{seo2021}. MEPOL directly computes a policy gradient based on the entropy estimate, while the others are artificial curiosity approaches. APT, APS, and Proto-RL are focused on learning representations, while RE3 uses random representations and is the most similar to our method.

Sparse literature exists on density and entropy estimation using k-means. \cite{wong1980} proved the inverse cube relationship between the probability density function and k-means cluster diameter for one dimension in the limit $k \rightarrow \infty$. 
\cite{miller2003} proposed density estimation using Voronoi diagrams where every cluster has the same probability measure, under assumptions of uniform probability within clusters, but without providing proofs or considering the geometric implications of clusters having the same probability measure. To our knowledge, no proofs for density estimation in more than one dimension using k-means were previously provided in the literature.
\section{Background}
\paragraph{Reinforcement learning.}
An RL problem can be formalized as a Markov Decision Process (MDP) defined by a 6-tuple $(\mathcal{S}, \mathcal{A}, T, R, s_{0}, \gamma)$. The state space $\mathcal{S}$ determines the possible states of the environment and the action space $\mathcal{A}$ the possible actions of the agent. The agent learns a policy function $\pi : \mathcal{S} \mapsto \mathcal{A}$ mapping states to actions at every time step. The environment starts in state $s_{0}$ and transitions to a state as a function of the previous state and previous agent action, according to the transition function $T : \mathcal{S} \times \mathcal{A} \mapsto \mathcal{S}$. 
The agent receives an extrinsic reward $r \in \mathbb{R}$ given by the reward function $R : \mathcal{S} \times \mathcal{A} \mapsto \mathbb{R}$.  The optimal agent maximizes the discounted cumulative extrinsic reward or return: $\pi^{*} \triangleq \argmax_{\pi}\sum_{t=0}^{\infty}\gamma^{t}r_{t}$. The discount factor $\gamma \in (0,1)$ determines the time horizon of the optimal agent and ensures that the return is finite if the reward function is bounded.
\paragraph{State visitation distribution.} The agent-environment interaction induces a state visitation distribution $p_{\pi}(s) \triangleq (1-\gamma)\sum_{t=0}^{\infty}\gamma^{t}p(s_{t}=s | \pi)$, where $p(s_{t}=s | \pi)$ is the probability of being in state $s$ at time step $t$ under policy $\pi$. The probability $p(s_{t}=s|\pi)$ can be computed from the policy, transition function and starting state. However, we generally do not have an analytical form for the transition function and can only sample from it through interactions with the environment. Intuitively, most information relevant to exploration is captured by the state visitation distribution.
\paragraph{Artificial curiosity.} Artificial curiosity algorithms generate intrinsic rewards $r_{i}$ at every time step. The agent then maximizes an augmented reward $\hat{r} = r + \beta r_{i}$ where $\beta \in \mathbb{R}_{+}$ is the reward scaling. \cite{schmidhuber1991} formulated the intrinsic reward as the state prediction error of a learned model of the transition function. Many approaches have since emerged based on state prediction error \citep{pathak2017}, state visitation counts \citep{tang2016}, state novelty \citep{burda2018}, ensemble disagreement \citep{pathak2019} and mutual information \citep{houthooft2016}, among others.
\paragraph{Maximum entropy exploration.} Maximum entropy exploration uses the differential entropy $H(p) \triangleq -\int p(x)\log p(x) \mathrm{d}x$ of the state visitation distribution $p_{\pi}$ as a maximization objective for exploration. 
Without additional constraints, the maximum entropy probability distribution is the uniform distribution. Thus, maximum entropy exploration drives the agent towards visiting environment states uniformly. 
Entropy can be maximized using an artificial curiosity approach, using the total entropy or change in entropy as an intrinsic reward. As the intrinsic reward must be computed at every time step, the computation of entropy must be efficient. 
\paragraph{k-Means clustering.} k-Means is a clustering algorithm that separates points into $k$ clusters. Iteratively, points are assigned to their closest cluster center and the cluster centers are updated to the average of their assigned points. 
In online k-means, a point $x$ is assigned to its closest cluster $\argmin_{i}\|\mu_{i} -  x\|$ and that cluster center is then updated according to $\mu_{i} \leftarrow \alpha x + (1 - \alpha)\mu_{i}$ where $\alpha \in (0,1)$ is the learning rate. 
\paragraph{Voronoi diagrams.}
The decision regions of a k-means clustering form a Voronoi diagram, which partitions the space into clusters $c_{i} = \{ x : \; \forall j \neq i . \; \| \mu_{i} -  x \| \leq \| \mu_{j}  -  x \| \}$. 
In an additively-weighted Voronoi diagram, also known as a hyperbolic Dirichlet tesselation, cluster assignments additionally depend on weights subtracted from the distances, and the decision regions become $c_{i} = \{ x : \; \forall j \neq i . \; \| \mu_{i} -  x \| - w_{i} \leq \| \mu_{j}  -  x \| - w_{j} \}$. Unlike in regular Voronoi diagrams, the decision surfaces of additively-weighted Voronoi diagrams are not straight lines at the midpoint between cluster centers, but rather hyperbolas with cluster centers as foci, and consequently the clusters are star-shaped instead of convex. 
\section{Approach}
We derive an optimization objective for maximizing a lower bound on the approximate entropy of the state visitation distribution under idealized conditions. Then, we construct a practical algorithm for efficiently generating intrinsic rewards based on the objective. 
Proofs are given in the appendix. 
\paragraph{Preliminaries.} Let $(\mathcal{S}, \mathcal{F})$ be a measure space where $\mathcal{S} \subset \mathbb{R}^{d}$ is a bounded subset of $d$-dimensional Euclidean space. On $\mathcal{F}$, let $m$ be the Lebesque measure, $P$ a probability measure and $p \triangleq \frac{\mathrm{d}P}{\mathrm{d}m}$ a continuous probability density function with support $\mathcal{X}$. 
\subsection{Density estimation using k-means}
Voronoi diagrams of k-means clusterings contain information about the probability density function that the clustered points were sampled from. Intuitively, smaller clusters occur in higher probability regions. This insight has only been proven for one dimension as an inverse cube relationship between the probability density function and cluster diameter \citep{wong1980}. To get a similar result for arbitrary dimensions, we use a variant of k-means where every cluster contains the same probability measure:
\paragraph{Definition 1.} A balanced Voronoi diagram is an additively-weighted Voronoi diagram where the weights $w$ are chosen such that $\forall i . \; P(c_{i}) = \int_{c_{i}}p(x)\mathrm{d}m(x) = \frac{1}{k}$.\\\\
In the finite case, the balanced property becomes that every cluster contains the same number of points. We are now ready to state our density estimation result:
\paragraph{Theorem 1.} Let $(V_{k})_{k=1}^{\infty}$ be a sequence of balanced Voronoi diagrams where $V_{k}$ has cluster centers $(\mu_{1},...,\mu_{k})$ dense in $\mathcal{X}$ as $k \rightarrow \infty$ and weights $(w_{1},...,w_{k})$ satisfying $\exists \eta \in (0,1). \; \forall k. \; \forall i,j . \; | w_{i} - w_{j} | \leq \eta\|\mu_{i} - \mu_{j} \|$. For any $x\in\mathcal{X}$, let $c^{k}_{i}$ be cluster in $V_{k}$ that $x$ belongs to. Then,
\begin{align*}
\lim\limits_{k\to\infty}\frac{1}{km(c^{k}_{i})} = p(x)
\end{align*}
(Proof is given in Appendix \ref{prf:thm_1}). This result says that the probability density at a point in cluster $c_{i}$ can be approximated as inversely proportional to the cluster measure $m(c_{i})$ for a large enough number of clusters $k$. This is a powerful result for non-parametric density estimation in arbitrary dimensions. However, the cluster measure $m(c_{i})$ is computationally intractible in general.
\subsection{Approximate entropy lower bound}
The entropy estimate obtained by substituting $p(x) \approx \frac{1}{km(c_{i})}$ in the definition of entropy is intractible, so we instead maximize a lower bound on the entropy estimate. To obtain such a bound, we use the insight that the boundary of a ball with center $\mu_{i}$ inscribed in cluster $c_{i}$ touches the decision boundary between the cluster and its closest neighbor. We obtain an intractible approximation of entropy using Theorem 1 and then derive a tractible lower bound on the approximation:
\paragraph{Theorem 2.} Let $\mu$ be the cluster centers and $w$ the weights of a balanced Voronoi diagram in $d$ dimensions, then for a sufficiently large number of clusters $k$,
\begin{align*}
  H(p) \gtrapprox \frac{d}{k}\sum_{i=1}^{k}\log (\min_{j \neq i} \| \mu_{i}  -  \mu_{j} \| + w_{i} - w_{j}) + \log \frac{\pi^{d/2}}{\Gamma(\frac{d}{2}+1)} - d
\end{align*}
where $\Gamma$ is Euler's gamma function. (Proof is given Appendix \ref{prf:thm_2}). Tighter bounds based on balls centered at cluster centers are not possible. The result shows that we maximize an approximate lower bound on the entropy $H(p)$ by maximizing the objective 
\begin{equation}
  L(\mu, w) = \sum_{i=1}^{k}\log (\min_{j \neq i} \| \mu_{i}  -  \mu_{j} \| + w_{i} - w_{j} )
  \label{eqn:objective}
\end{equation}
since $d$, $k$ and $\frac{\pi^{d/2}}{\Gamma(\frac{d}{2}+1)}$ are constants that do not depend on the clustering. Intuitively, the clusters are a summary of the agent's trajectories, and the objective is maximized when the clusters are spread out as much and evenly as possible; this is consistent with the idea of visiting states uniformly.
\subsection{Algorithm}
\paragraph{Intrinsic reward.} We optimize the objective in Equation \ref{eqn:objective} using an artificial curiosity approach, generating an intrinsic reward as the difference in the objective before and after encountering a state: 
\begin{equation}
  r_{i} = L(\tilde{\mu}, \tilde{w}) - L(\mu, w)
  \label{eqn:reward}
\end{equation}
where $\tilde{\mu}$ and $\tilde{w}$ are the centers and weights after updating our clustering based on state $s$.
In practice, we use a generalized optimization objective to avoid numerical issues due to the logarithm function in Equation \ref{eqn:objective} when distances between cluster centers are small:
\begin{equation}
  L_{f}(\mu,w) = \sum_{i=1}^{k} f(\min_{j \neq i} \| \mu_{i}  -  \mu_{j} \| + w_{i} - w_{j})
\label{eqn:practical_objective}
\end{equation}
where the function $f : \mathbb{R} \mapsto \mathbb{R}$ satisfies $\argmax_{x}\sum_{i=1}^{k}f(x_{i})=\argmax_{x}\sum_{i=1}^{k}\log x_{i}$ for bounded $\sum_{i=1}^{k}x_{i}$ so the optimum is unchanged.
We found $f(x)=\sqrt{x}$ to have good empirical performance.
\paragraph{Clustering.} For clustering, we use additively-weighted online k-means, where weights $w$ are subtracted from distances during cluster assignments. A state $s$ is assigned to its closest cluster 
\begin{equation}
  \argmin_{i}\|\mu_{i} -  s\| - w_{i}
\end{equation}
and that cluster center is then updated according to
\begin{equation}
  \mu_{i} = \alpha x + (1 - \alpha)\mu_{i} 
\end{equation}
where $\alpha \in (0,1)$ is the learning rate. The weights are computed by maintaining a count $n$ of the number of points assigned to every cluster and considering the standard deviation:
\begin{equation}
  w_{i} = \kappa(\frac{1}{k}\sum_{j=1}^{k}n_{j}-n_{i}) 
\end{equation}
where $\kappa \in \mathbb{R_{+}}$ is the balancing strength. 
\\\\We found cluster stability issues to occur in practice due to the use of online k-means coupled with correlation between neighboring states in agent trajectories. To improve stability in on-policy RL, we perform batch updates on shuffled trajectories every $B$ time steps, computing rewards without updating clusterings in between. The stabilizing effect of batch updates makes our approach promising for distributed on-policy RL. Pseudocode for our reward computation is shown in Algorithm \ref{alg:reward} and on-policy RL using our approach is shown in Algorithm \ref{alg:rl} (Appendix \ref{apx:exploration}).
\begin{algorithm}
\caption{Intrinsic reward computation of KME.}
  \begin{algorithmic}
    \State \textbf{hyperparameters}
    \State $k : \text{Number of clusters}$
    \State $\alpha : \text{Learning rate}$
    \State $\kappa : \text{Balancing strength}$
    \State \textbf{variables} \Comment{Space complexity}
    \State $\mu: \text{Cluster centers}$ \Comment{$\mathcal{O}(kd)$}
    \State $n: \text{Cluster counts}$ \Comment{$\mathcal{O}(k)$}
    \State $m: \text{Closest clusters} \quad \argmin_{j \neq i}\|\mu_{i} -  \mu_{j}\| + \kappa(n_{j} - n_{i})$ \Comment{$\mathcal{O}(k)$}
    \State $M: \text{Closest distances} \quad\ \ \; \min_{j \neq i}\|\mu_{i} -  \mu_{j}\| + \kappa(n_{j} - n_{i})$ \Comment{$\mathcal{O}(k)$}
    \Function{ComputeReward}{s} \Comment{Time complexity}
    \State $L_{1} \gets \sum_{i=1}^{k}f(M_{i})$ \Comment{$\mathcal{O}(k)$}
    \State $i^{*} \gets \argmin_{i}\|\mu_{i} -  s\| - \kappa(\frac{1}{k}\sum_{j=1}^{k}n_{j}-n_{i})$ \Comment{$\mathcal{O}(kd)$}
    \State $\mu_{i^{*}} \gets \alpha s + (1 - \alpha)\mu_{i^{*}}$ \Comment{$\mathcal{O}(d)$}
    \State $n_{i^{*}} \gets n_{i^{*}} + 1$ \Comment{$\mathcal{O}(k)$}
    \State $m,M \gets \textsc{UpdateClosestDistances($i^{*}$)}$ \Comment{$\mathcal{O}(k^{2}d)^{*}$}
    \State $L_{2} \gets \sum_{i=1}^{k}f(M_{i})$ \Comment{$\mathcal{O}(k)$}
    \State \textbf{return} $L_{2} - L_{1}$
    \EndFunction
  \end{algorithmic}
  \label{alg:reward}
\end{algorithm}
\paragraph{Hyperparameters.} The effects of hyperparameters are decoupled but complex. Increasing $B$ improves stability, but delays the informativeness of rewards. 
Increasing $\kappa$ reduces stability, but ensures that the Voronoi diagram is balanced and subsequently improves the accuracy of the entropy estimate.
$\alpha$ similarly reduces stability when increased, but ensures that the clustering is representative of the agent trajectories.  
Increasing $k$ improves stability and the accuracy of the entropy estimate, however it is the sole hyperparameter affecting the computational complexity. 
\paragraph{Computational complexity.} The reward in Equation \ref{eqn:reward} can be computed efficiently by maintaining an array with the closest neighbor of each cluster. The reward computation then has space complexity $\mathcal{O}(kd)$ and time complexity $\mathcal{O}(k^{2}d)$, as seen from Algorithm \ref{alg:reward}. 
However, this worst case time complexity stems from the pathological case where the updated cluster is the closest neighbor of all other clusters before an update and none after. The practical time complexity is $\mathcal{O}(kd)$.
\section{Experiments}
\begin{figure}
  \centering
  \includegraphics[width=0.16\textwidth]{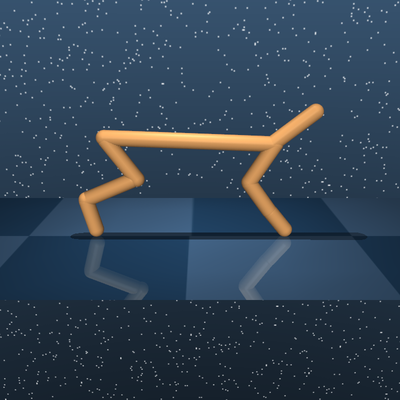}
  \includegraphics[width=0.16\textwidth]{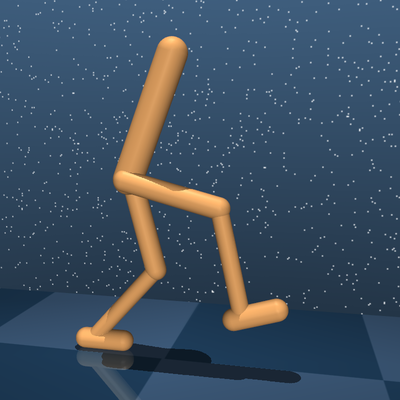}
  \includegraphics[trim=0 0 0 42, clip, width=0.16\textwidth]{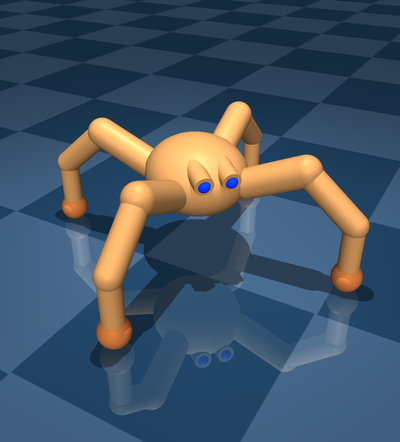}
  \includegraphics[width=0.16\textwidth]{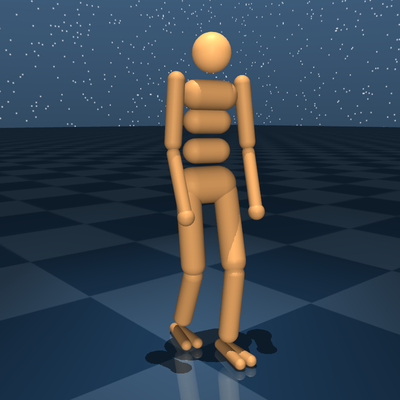}
  \includegraphics[width=0.16\textwidth]{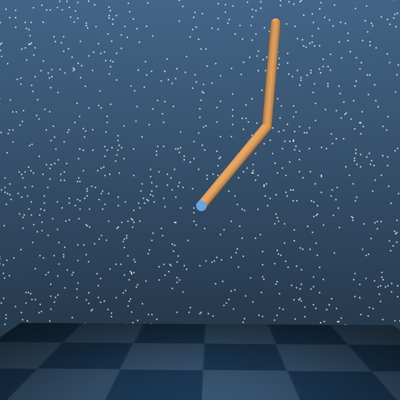}
  \includegraphics[width=0.16\textwidth]{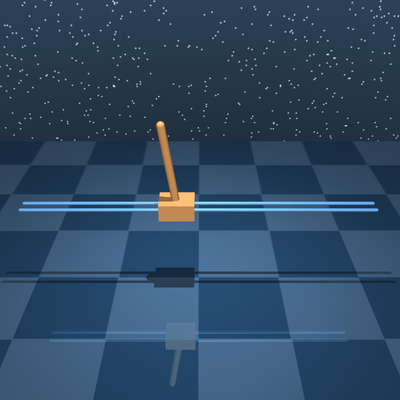}
  \caption{MuJoCo tasks. Left to right: Cheetah, Walker, Quadruped, Humanoid, Acrobot, Cartpole. The agent must swing a pole up in Cartpole and Acrobot, and move above a speed in the other tasks. Cheetah, Walker, Quadruped and Humanoid were adapted to make rewards sparse. The state and actions spaces range from $(\mathbb{R}^{5}, \mathbb{R}^{1})$ in Cartpole to $(\mathbb{R}^{67},\mathbb{R}^{21})$ in Humanoid.} 
  \label{fig:mujoco_domains}
\end{figure}

\begin{figure}
  \centering
  \includegraphics[trim=25 25 50 65,clip, width=0.49\textwidth]{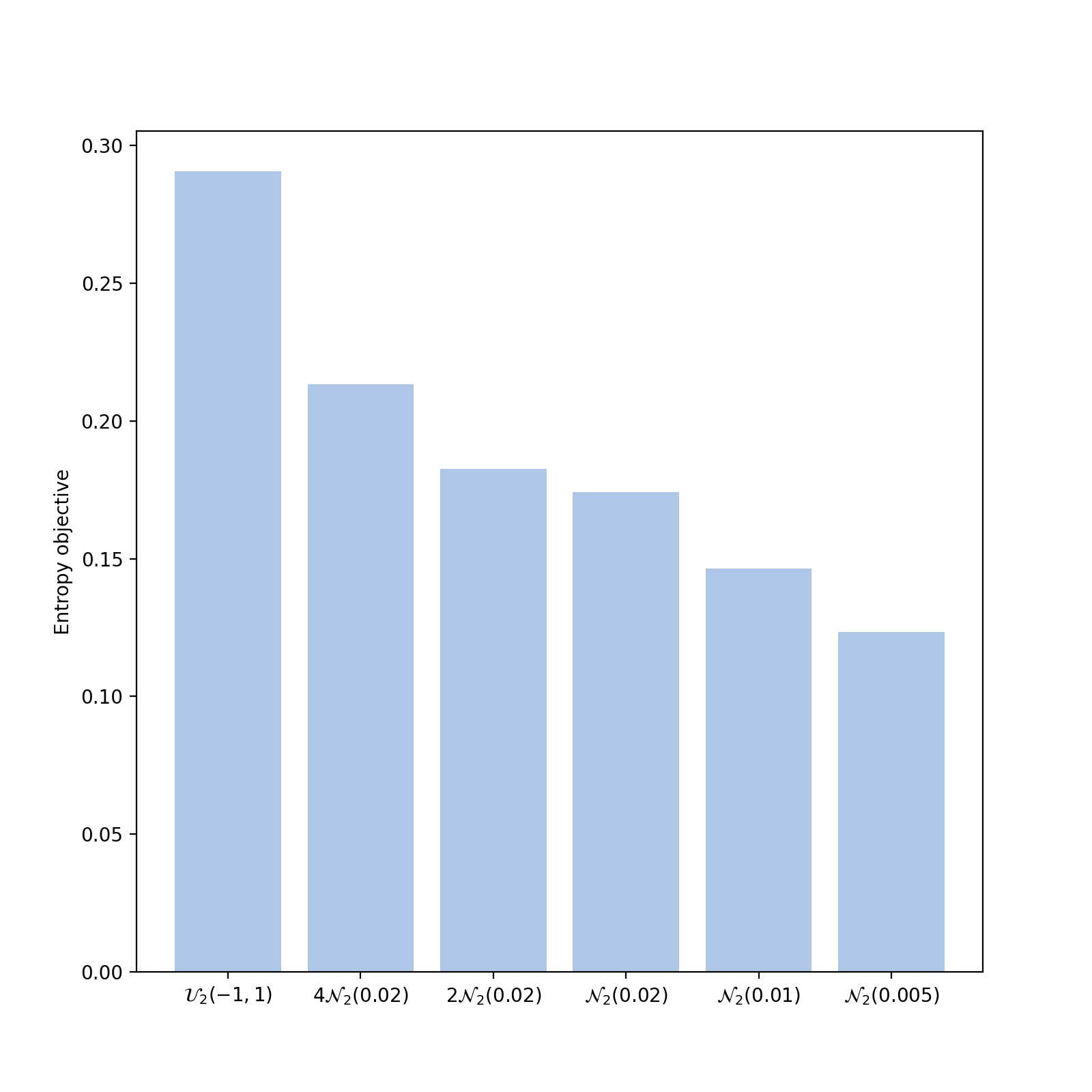}
  \includegraphics[trim=25 25 50 65,clip, width=0.49\textwidth]{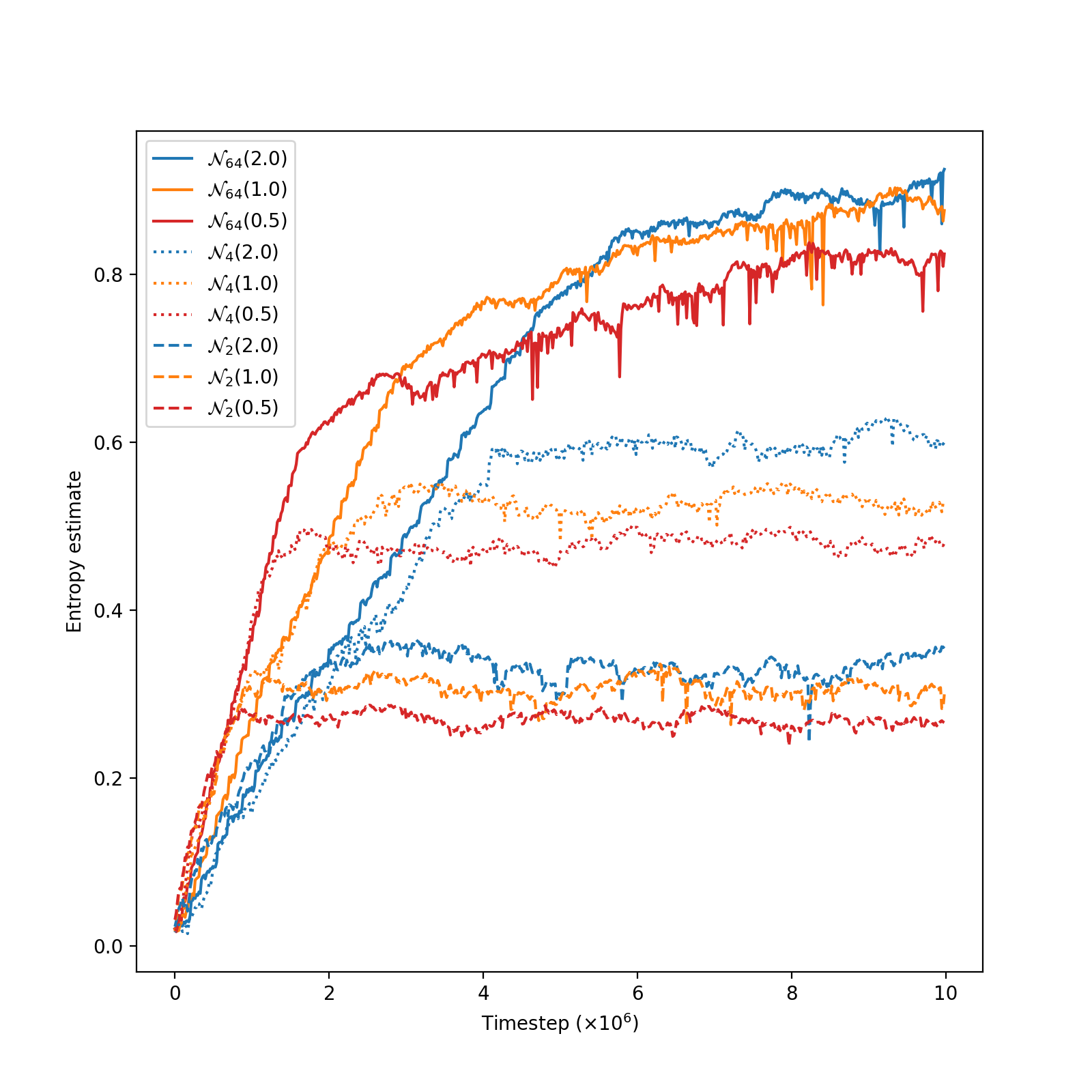}
  \caption{Entropy estimation results. 
  $\mathcal{U}_{d}$ denotes a uniform distribution in $d$ dimensions, $\mathcal{N}(\sigma^{2})$ a Gaussian with $\sigma^{2}$ diagonal covariance and $n\mathcal{N}$ a mixture of $n$ Gaussians. 
  \textbf{Left}: Random samples from distributions in 2 dimensions. The entropy estimate preserves the decreasing ordering of entropy from left to right. 
  The points and clusters from this experiment are shown in Figure \ref{fig:entropy_distributions}. 
  \textbf{Right}: Random walks on Gaussians with different variances in 2, 4 and 64 dimensions. The entropy estimate also preserves the entropy ordering from blue to orange to red in each dimension, but converges slowly for higher dimensions and variances due to initializing cluster centers to zero. 
  }
  \label{fig:entropy_results}
\end{figure}

\begin{figure}
  \centering
  \includegraphics[width=0.16\textwidth]{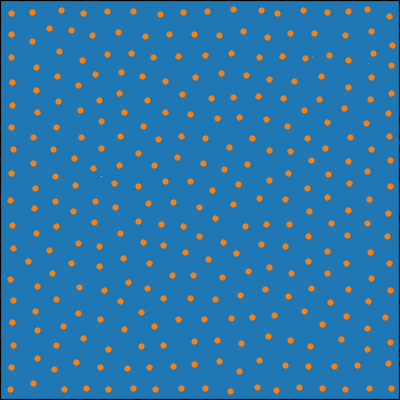}
  \includegraphics[width=0.16\textwidth]{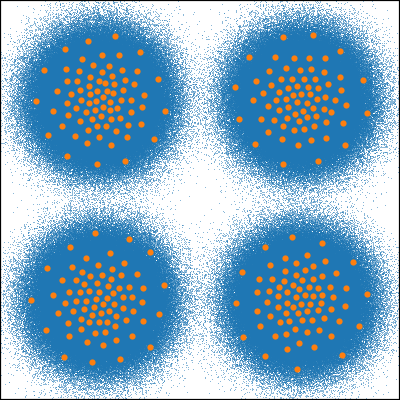}
  \includegraphics[width=0.16\textwidth]{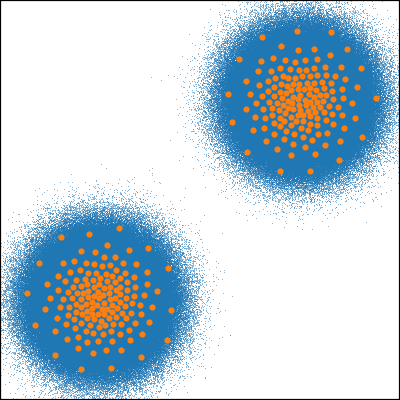}
  \includegraphics[width=0.16\textwidth]{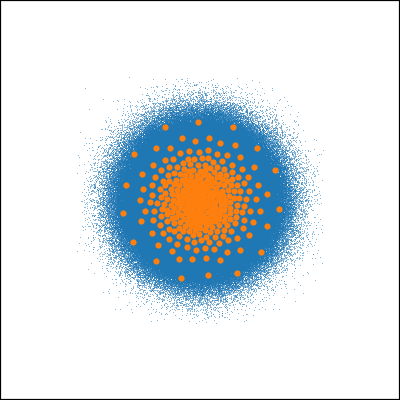}
  \includegraphics[width=0.16\textwidth]{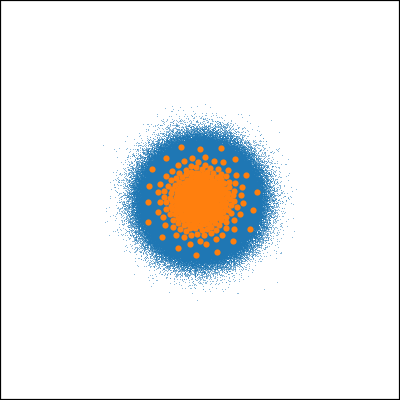}
  \includegraphics[width=0.16\textwidth]{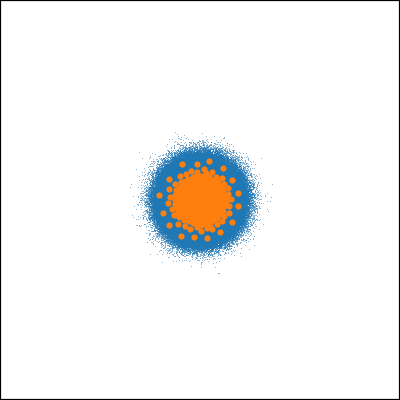}
  \caption{Points sampled from probability distributions (blue) with k-means cluster centers (orange). Left to right (decreasing entropy): Uniform, Gaussian mixtures, Gaussians. Observe that the distances between cluster centers contain information about the probability density function and entropy.}
  \label{fig:entropy_distributions}
\end{figure}

\paragraph{Exploration.}
We evaluated our algorithm on sparse reward MuJoCo tasks from DeepMind Control Suite \citep{tassa2020}. The tasks are standard benchmarks for RL in continuous spaces and are shown in Figure \ref{fig:mujoco_domains}. For RL, we trained neural networks using Proximal Policy Optimization (PPO) \citep{schulman2017} to maximize extrinsic rewards augmented by intrisic rewards from our method and baseline methods. 
For baselines, we used RND \citep{burda2018} as it is a standard exploration baseline and RE3 \citep{seo2021} as it is the most similar method to ours.
    We used the same hyperparameters across tasks. Details about the task and algorithm implementations are given in Appendix \ref{apx:exploration}.
The results in Figure \ref{fig:mujoco_results} show that our method is competitive on these benchmarks, especially on Cheetah and Quadruped where PPO is unable to find rewards.
\paragraph{Entropy estimation.} 
To investigate how meaningful our optimization objective from Equation \ref{eqn:practical_objective} is as an entropy estimate, we performed experiments to evaluate whether the objective preserves entropy orderings on (1) samples from different probability distributions and (2) samples that are high-dimensional and correlated like states from agent trajectories in high-dimensional RL. For (1) we sampled points from different probability distributions in 2 dimensions. 
The sampled points and k-means clusterings from are shown in Figure \ref{fig:entropy_distributions}. 
For (2) we sampled points from random walks on Gaussians with different variances in different dimensions.
We used the same hyperparameters from the MuJoCo tasks. 
From Figure \ref{fig:entropy_results}, we see that the objective preserves entropy orderings even for random walks in very high (64) dimensions. This suggests that it is meaningful as an entropy estimate for the state visitation distribution in high-dimensional, continuous RL.
\paragraph{Time complexity.} For both the exploration and entropy estimation experiments, we recorded the average number of pathological updates in Algorithm \ref{alg:reward}. 
 We found these updates to occur only during a tiny fraction ($\approx 0.01\%$) of steps. Thus, the practical time complexity of our algorithm is linear in $k$.
\begin{figure}
  \centering
  \includegraphics[trim = 60 30 60 45, clip, width=\textwidth]{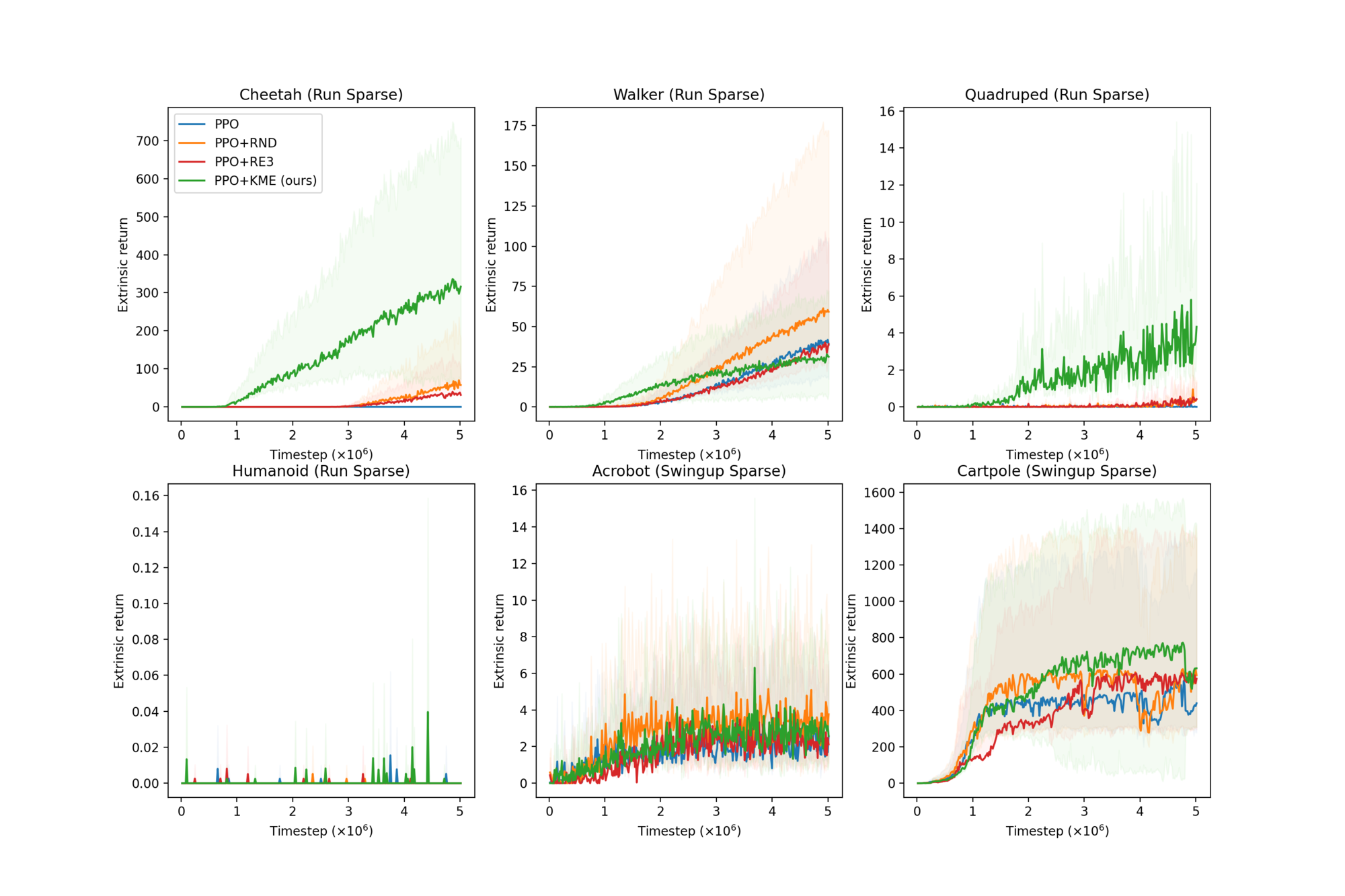}
  \caption{Performance on sparse MuJoCo tasks. The plots show mean extrinsic returns with 95\% confidence intervals over 5 random seeds. Higher is better. Our method (green) is competitive and outperforms RND (orange) and RE3 (red) on Cheetah and Quadruped where PPO (blue) is unable to find rewards. None of the methods make significant progress on Humanoid.}
  \label{fig:mujoco_results}
\end{figure}
\section{Discussion and conclusion}
Our theory is based on idealized conditions that may not be satisfied in practice. Namely, we assumed (1) continuity of the probability density function, (2) denseness of cluster centers in the limit $k \rightarrow \infty$ and (3) existence of cluster weights with $\| w_{i} - w_{j} \|  \leq \eta\| \mu_{i} - \mu_{j} \|$ such that $P(c_{i})=\frac{1}{k}$. These assumptions are sufficient conditions for the density estimate result in Theorem 1. There may exist weaker necessary conditions, especially for (2) and (3) which are used in our proof to show that the diameter of clusters can be made arbitrarily small; there exist counterexamples showing that (1) and $P(c_{i})=\frac{1}{k}$ are insufficient conditions for this. 
In practice, k-means clusterings where cluster diameters do not decrease with $k$ are unstable and tend not to occur. 
However, it is difficult to characterize the geometry of these algorithmically generated structures.
Despite the interesting geometric and information-theoretic properties of balanced Voronoi diagrams, they remain largely unstudied. The characterization of balanced Voronoi diagrams, especially those formed by k-means clustering algorithms, is a challenging but promising direction of research. 

For our practical algorithm, we assumed that our variant of k-means forms a balanced Voronoi diagram, used a limited number of clusters and replaced the logarithm in Equation \ref{eqn:objective} with a square root. However, our empirical results suggest that this practical approximation of theory is adequate for both exploration and entropy estimation. 
The algorithm did not manage to find rewards on Humanoid, which shows a limitation on very high-dimensional problems. An improvement to our method would be to use representation learning to instead estimate entropy in a lower-dimensional embedding space. Sucessor features \citep{dayan1993} are here especially promising as they improve the continuity of the state visitation distribution. The slow convergence of our algorithm during random walks in high dimensions occurs due to cluster centers being initialized to zero and rarely changing unless points are sampled near them. This is less of a problem for RL with episodic state resets (if the stating state is zero), but suggests that our method can be improved with more thoughtful initialization.

Our findings support the idea that maximum entropy is a promising theoretical foundation for exploration in continuous spaces.
Maximum entropy exploration is consistent with count-based exploration theory based on the \textit{optimism in the face of uncertainty} heuristic \citep{strehl2005}\citep{kolter2009}\citep{bellamare2016}, in that both suggest to reward the least visited state maximally and visit states uniformly without extrinsic rewards. 
However, count-based exploration theory concerns the exploration-exploitation trade-off under random reward functions, while maximum entropy considers exploration as an aspect of the state visitation distribution.
We find the latter perspective more pertinent to real-world exploration.
It is unfortunate that differential entropy does not inherit the invariance under change of variables and non-negativity of discrete entropy, while it does inherit the usual algorithmic challenges and unintuitive geometry of high-dimensional spaces. 
The search for theoretical foundations that capture the problem of exploration in continuous spaces is an important endeavour and a promising direction of research. 
\begin{ack}
The authors are particularly grateful to Ethan Palmiere and Pau Vilimelis Aceituno for the many discussions that inspired this work. We also thank Seijin Kobayashi, Alexander Meulemans, João Sacramento and Benjamin Grewe. The stimulating research environments of the Institute of Neuroinformatics in Zürich and the African Institute for Mathematical Sciences in Kigali are appreciated.

This work was funded by the Swiss National Science Foundation under grant 182539 (Neuromorphic Algorithms based on Relational Networks). The authors declare no conflicts of interest.
\end{ack}
\bibliography{neurips}
\section{Appendix}
\subsection{Experiments}
\begin{algorithm}
\caption{On-policy RL using KME.}
  \begin{algorithmic}
    \State \textbf{hyperparameters}
    \State $B : \text{Batch size}$
    \State $\beta : \text{Reward scaling}$
    \Function{Learn}{$\text{policy } \pi, \text{MDP } (\mathcal{S}, \mathcal{A}, T, R, s_{0}, \gamma), \text{time horizon } t_{max}$}
      \State $\text{state } s \gets s_{0}$
      \State $\text{time step } t \gets 0$
      \While {$t < t_{max}$}
        \State $\text{trajectory } \tau \gets \{\varnothing\}$
        \ForAll {$i \in \{1,...,B\} $}
          \State $\text{action } a \gets \pi(s)$ 
          \State $\text{extrinsic reward } r \gets R(s,a)$
          \State $s \gets T(s,a)$
          \State $\text{intrinsic reward } r_{i} \gets \textsc{ComputeReward}(s)$ \Comment{Algorithm \ref{alg:reward} without variable updates.}
          \State $\tau \gets \tau \cup \{(s,a,r+\beta r_{i})\}$
        \EndFor
        \State $\pi \gets \textsc{UpdatePolicy}(\pi, \tau, \gamma)$ \Comment{RL algorithm.}
        \ForAll {$ (s, a, \hat{r}) \in \tau$}
          \State $\textsc{ComputeReward}(s)$ \Comment{Algorithm \ref{alg:reward} with variable updates.}
        \EndFor
        \State $t \gets t + 1$
      \EndWhile
      \State \textbf{return} $\pi$
    \EndFunction
  \end{algorithmic}
  \label{alg:rl}
\end{algorithm}
\label{apx:exploration}
\paragraph{Environment.} As the DeepMind Control Suite \citep{tassa2020} implementations of Humanoid, Quadruped, Walker and Cheetah only provide dense rewards, we set rewards to zero below a threshold $\tau$. Following \cite{seo2021}, we used $\tau=0.5$ for Cheetah and Walker. We found $\tau=0.2$ for Humanoid and $\tau=0.7$ for Quadruped as thresholds where PPO would no longer find rewards. The state and actions spaces ($\mathcal{S}, \mathcal{A}$) for the tasks are: Cheetah ($\mathbb{R}^{17}, \mathbb{R}^{6}$), Walker ($\mathbb{R}^{24}, \mathbb{R}^{6}$), Humanoid ($\mathbb{R}^{67}, \mathbb{R}^{21}$), Quadruped ($\mathbb{R}^{58}, \mathbb{R}^{12}$), Cartpole ($\mathbb{R}^{1}, \mathbb{R}^{5}$), Acrobot ($\mathbb{R}^{1}, \mathbb{R}^{6}$).
\paragraph{Algorithms.} For PPO, we used the Stable Baselines3 \citep{stable-baselines3} implementation with a rollout size $B=16384$ and 16 environments running in parallel. The default implementation uses neural networks with 2 hidden layers of size 64 and tanh activation functions for both the policy and value networks. Pseudocode for on-policy RL using KME is shown in Algorithm \ref{alg:rl}. We adapted the RE3 implementation from \citet{seo2021} and RND implementation from \cite{lashkin2021}. For both RE3 and RND, we used a neural network encoder with 2 hidden layers of size 1024 and ReLU activation functions. For RND, we used learning rate $\alpha=0.0001$, encoder output size $h=512$ and reward scaling $\beta=0.00001$. For RE3, we used $k=3$, $h=50$ and $\beta=0.0001$. For KME, we used $k=300$, $\alpha=0.05$, $\kappa=0.0001$ and $\beta=0.01$. Hyperparameters were found through parameter sweeps using WandB \citep{wandb} when default values were not used.
\subsection{Proofs}
\subsubsection{Proof of Theorem 1}
\label{prf:thm_1}
\paragraph{Theorem 1.} Let $(V_{k})_{k=1}^{\infty}$ be a sequence of balanced Voronoi diagrams where $V_{k}$ has cluster centers $(\mu_{1},...,\mu_{k})$ dense in $\mathcal{X}$ as $k \rightarrow \infty$ and weights $(w_{1},...,w_{k})$ satisfying $\exists \eta \in (0,1). \; \forall k. \; \forall i,j . \; | w_{i} - w_{j} | \leq \eta\|\mu_{i} - \mu_{j} \|$. For any $x\in\mathcal{X}$, let $c^{k}_{i}$ be cluster in $V_{k}$ that $x$ belongs to. Then,
\begin{align*}
\lim\limits_{k\to\infty}\frac{1}{km(c^{k}_{i})} = p(x)
\end{align*}
Note: By $(\mu_1,...,\mu_k)$ dense in $\mathcal{X}$ as $k\to\infty$, we mean in the uniform sense $\forall \delta > 0. \; \exists k_0. \; \forall x\in\mathcal{X}. \; \forall k > k_0 . \; \exists \mu \in (\mu_1,...,\mu_k).   \; ||x-\mu|| < \delta$. \\\\
Proof. For any $\hat{x} \in \mathcal{X}$, let $c_{i} \triangleq c_{i}^{k}$ be the cluster that $\hat{x}$ belongs to in $V_{k}$. We first show that there exists a point $x_{i} \in c_{i}$ such that $p(x_{i})=\frac{1}{km(c_{i})}$. To do this, we define:
  \begin{align*}
    \underline{x_{i}} \triangleq \argmin_{x \in \overline{c_{i}}}p(x) \\
    \overline{x_{i}} \triangleq \argmax_{x \in \overline{c_{i}}}p(x) \\
    p_{i} \triangleq \frac{\int_{c_{i}}p(x)\mathrm{d}m(x)}{\int_{c_{i}}\mathrm{d}m(x)} = \frac{1}{k m(c_{i})}
  \end{align*}
  where $\overline{c_{i}}$ denotes the closure of $c_{i}$. It then follows:
  \begin{align*}
    \int_{c_{i}}p(\underline{x_{i}})\mathrm{d}m(x) \leq \int_{c_{i}}p(x)\mathrm{d}m(x) \leq \int_{c_{i}}p(\overline{x_{i}})\mathrm{d}m(x) \Leftrightarrow p(\underline{x_{i}}) \leq p_{i} \leq p(\overline{x_{i}})
  \end{align*}
  Since $c_{i}$ is star-shaped (its intersection with any line through $\mu_{i}$ is convex), we can define a continuous function $\phi : [0,1] \rightarrow c_{i} \cup \{\underline{x_{i}}, \overline{x_{i}} \}$ with $\phi(0) = \underline{x_{i}}$ and $\phi(1) = \overline{x_{i}}$ as follows:
  \begin{align*}
    \phi(\xi) =
    \begin{cases}
      2\xi\mu_{i} + (1-2\xi)\underline{x_{i}}, & 0 \leq \xi \leq \frac{1}{2} \\
      (2\xi-1)\overline{x_{i}} + (2-2\xi)\mu_{i}, & \frac{1}{2} < \xi \leq 1
    \end{cases}
  \end{align*}
  By continuity of $p$, $\psi \triangleq p \circ \phi$ is continuous. Then, due to the intermediate value theorem, there exists a point $x_{i} \in c_{i}$ such that $p(x_{i}) = \frac{1}{k m(c_{i})}$. \\\\
  We now show that the diameter of cluster $c_{i}$ becomes arbitrarily small as $k$ gets large:
  \begin{align*}
    \forall \delta > 0. \; \exists k_{0} . \; \forall k > k_{0}. \; \max_{x \in c_{i}} \| \mu_{i}  -  x \| < \delta
  \end{align*}
We claim that this holds for any cluster, which implies the statement. Assume for contradiction that this is not the case. Let $x$ be an arbitrary point in $\mathcal{X}$, $\mu$ the center of the cluster that $x$ belongs to in $V_k$, and $\mu_\epsilon$ some different cluster center in $V_k$ with $\| \mu_e - x \| = \epsilon$. For some $\delta > 0$, every $k_0$ and some $k > k_0$, we have $\| \mu - x \| \geq \delta$ by assumption. However,
\begin{align*}
  \delta&\leq \| \mu - x \| \\
  &\leq \| \mu_\epsilon - x \| - w_\epsilon + w  \\
  &\leq \| \mu_\epsilon - x \| + |w - w_\epsilon| \\
  &\leq \| \mu_\epsilon - x \| + \eta\| \mu - \mu_\epsilon \| \\ 
  &\leq (1 + \eta)\epsilon + \eta\| \mu - x\| 
\end{align*}
where the last inequality follows from the triangle inequality.
Since $(\mu_{1},...,\mu_{k})$ is dense in $\mathcal{X}$ as $k \rightarrow \infty$, we can obtain the contradiction by setting $\epsilon < \frac{1-\eta}{1+\eta}\delta$ as this implies 
\begin{align*}
 \| \mu - x \| \leq \frac{1+\eta}{1-\eta}\epsilon < \delta
\end{align*}
  Now, since $p$ is continuous and $\| x_{i} - x \| \leq 2\max_{y \in c_{i}} \| \mu_{i} - y \|$ for all $x \in c_{i}$:
  \begin{align*}
    \forall \varepsilon > 0. \;\exists k_{0} . \; \forall k > k_{0}. \; | \frac{1}{k m(c_{i})}  -  p(\hat{x}) | < \varepsilon
  \end{align*}
  Since $\hat{x}$ was arbitrary, this concludes the proof. $\square$
\subsubsection{Proof of Theorem 2}
\label{prf:thm_2}
\paragraph{Theorem 2.} Let $\mu$ be the cluster centers and $w$ the weights of a balanced Voronoi diagram in $d$ dimensions, then for a sufficiently large number of clusters $k$,
\begin{align*}
  H(p) \gtrapprox \frac{d}{k}\sum_{i=1}^{k}\log (\min_{j \neq i} \| \mu_{i}  -  \mu_{j} \| + w_{i} - w_{j}) + \log \frac{\pi^{d/2}}{\Gamma(\frac{d}{2}+1)} - d
\end{align*}
Proof. For neighboring clusters, denote a segment between them and their boundary by
  \begin{align*}
    l_{ij} \triangleq \{x \in \mathcal{X} : \lambda_{ij}\mu_{j} + (1 - \lambda_{ij})\mu_{i}, \; \lambda_{ij} \in [0,1] \}  \\
  b_{ij} \triangleq \{x \in \mathcal{X} : \| \mu_{i} - x \| - \| \mu_{j}  -  x \| = w_{i} - w_{j} \}
  \end{align*}
  Then, $\lambda_{ij}$ at their intersection is given by:
  \begin{align*}
    \|\mu_{i} -  (\lambda_{ij}\mu_{j} + (1 - \lambda_{ij})\mu_{i})\|) - \|\mu_{j} - (\lambda_{ij}\mu_{j} + (1- \lambda_{ij})\mu_{i})\|= w_{i} - w_{j} \Leftrightarrow \lambda_{ij} = \frac{1}{2} + \frac{w_{i} - w_{j}}{2\|\mu_{i} -  \mu_{j}\|}) 
  \end{align*}
We claim that $\mathcal{B}_{\min_{j\neq i}\lambda_{ij}\|\mu_{i}-\mu_{j}\|}(\mu_{i})\subseteq c_{i}$. Assume $w_{i} < w_{j}$ as the statement is trivial otherwise since the decision boundary is a straight line or a hyperbola closer to focus $\mu_{j}$ if $w_{i} \geq w_{j}$. The decision boundary is then a rotationally symmetric hyperbola closer to focus $\mu_{i}$. Since the curvature of a hyperbola closer to focus $\mu_{i}$ is upper bounded by the curvature of a sphere with center $\mu_{i}$, we get the inclusion. \\\\
  The bound then follows from
  \begin{align*}
    H(p) &\triangleq -\int p(x) \log p(x) \mathrm{d}m(x) \\
    &\overset{(1)}{=} -\sum_{i=1}^{k} \int_{c_{i}}p(x) \log p(x) \mathrm{d}m(x) \\
    &\overset{(2)}{\approx} -\sum_{i=1}^{k} \int_{c_{i}}p(x) \log \frac{1}{km(c_{i})} \mathrm{d}m(x) \\
    &\overset{(3)}{=}\log k + \frac{1}{k}\sum_{i=1}^{k} \log m(c_{i}) \\
    &=\log k + \frac{1}{k}\sum_{i=1}^{k} \log \frac{m(c_{i})}{m(\mathcal{B}_{\min_{j \neq i} \lambda_{ij} \| \mu_{i}  -  \mu_{j} \|})} +  \frac{1}{k}\sum_{i=1}^{k} \log m(\mathcal{B}_{\min_{j \neq i} \lambda_{ij} \| \mu_{i}  -  \mu_{j} \|}) \\ 
    &\overset{(4)}{=}\log k + \frac{1}{k}\sum_{i=1}^{k} \log \frac{m(c_{i})}{m(\mathcal{B}_{\min_{j \neq i}\lambda_{ij} \| \mu_{i}  -  \mu_{j} \|})} +  \frac{1}{k}\sum_{i=1}^{k} \log \frac{\pi^{d/2}}{\Gamma(\frac{d}{2}+1)}(\min_{j \neq i} (\frac{1}{2} + \frac{w_{i}-w_{j}}{2\|\mu_{i} - \mu_{j}\|}) \| \mu_{i}  -  \mu_{j} \|)^{d} \\ 
    &=\underbrace{\log k}_{\geq 0} + \underbrace{\frac{1}{k}\sum_{i=1}^{k} \log \frac{m(c_{i})}{m(\mathcal{B}_{\min_{j \neq i}\lambda_{ij} \| \mu_{i}  -  \mu_{j} \|})}}_{\geq 0} + \log \frac{\pi^{d/2}}{\Gamma(\frac{d}{2}+1)} - d +  \frac{d}{k}\sum_{i=1}^{k} \log (\min_{j \neq i} \| \mu_{i}  -  \mu_{j}\| + w_{i} - w_{j}) \\
    &\overset{(5)}{\geq} \frac{d}{k}\sum_{i=1}^{k}\log(\min_{j \neq i} \| \mu_{i}  -  \mu_{j} \| + w_{i} - w_{j}) + \log \frac{\pi^{d/2}}{\Gamma(\frac{d}{2}+1)} - d
  \end{align*}
  where (1) follows as $(c_{1},...,c_{k})$ is a partition of $\mathcal{X}$, (2) follows from Theorem 1 for large $k$, (3) follows from $P(c_{i})=\frac{1}{k}$, (4) follows from the Lebesque measure of a ball in $d$-dimensional Euclidean space and (5) follows from $\mathcal{B}_{\min_{j\neq i}\lambda_{ij}\|\mu_{i}-\mu_{j}\|}(\mu_{i})\subseteq c_{i}$ and $k \geq 1$. $\square$

\end{document}